\documentclass[10pt,twocolumn,letterpaper]{article}

\usepackage{cvpr}
\usepackage{times}
\usepackage{epsfig}
\usepackage{graphicx}
\usepackage{amsmath}
\usepackage{amssymb}
\usepackage{subfigure}
\usepackage{algorithm}
\usepackage{algorithmic}
\newcommand{\argmin}{\mathop{\rm arg~min}\limits}
\newcommand{\argmax}{\mathop{\rm arg~max}\limits}

\usepackage[pagebackref=true,breaklinks=true,letterpaper=true,colorlinks,bookmarks=false]{hyperref}

\cvprfinalcopy 

\begin{document}

\title{DeMIAN: Deep Modality Invariant Adversarial Network}
\author{Kuniaki Saito, Yusuke Mukuta, Yoshitaka Ushiku, Tatsuya Harada\\
The University of Tokyo\\
\and
{\tt\small \{k-saito, mukuta, ushiku, harada\}@mi.t.u-tokyo.ac.jp}
}

\maketitle

\begin{abstract}
Obtaining common representations from different modalities is important in that they are interchangeable with each other in a classification problem. For example, we can train a classifier on image features in the common representations and apply it to the testing of the text features in the representations. Existing multi-modal representation learning methods mainly aim to extract rich information from paired samples and train a classifier by the corresponding labels; however, collecting paired samples and their labels simultaneously involves high labor costs. Addressing paired modal samples without their labels and single modal data with their labels independently is much easier than addressing labeled multi-modal data. To obtain the common representations under such a situation, we propose to make the distributions over different modalities similar in the learned representations, namely modality-invariant representations. In particular, we propose a novel algorithm for modality-invariant representation learning, named Deep Modality Invariant Adversarial Network (DeMIAN), which utilizes the idea of Domain Adaptation (DA). 
Using the modality-invariant representations learned by DeMIAN, we achieved better classification accuracy than with the state-of-the-art methods, especially for some benchmark datasets of zero-shot learning.
\end{abstract}
 \begin{figure}[t!]
  \begin{center}
   \includegraphics[width=\hsize]{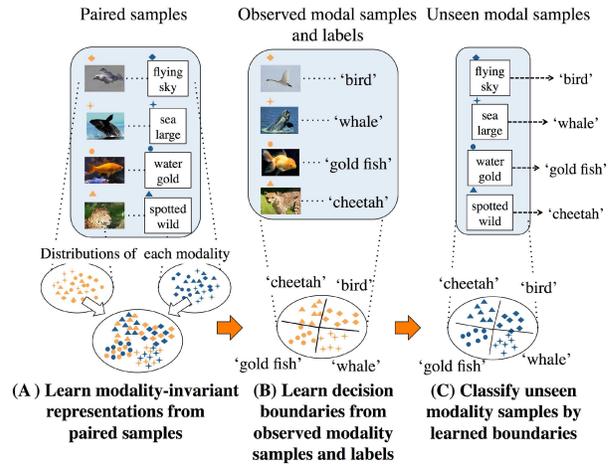}
  \end{center}
  \vspace{-2mm}
    \caption{Illustration of  our proposed method. We proposed a method that learns modality-invariant representations. (A) We aim to learn modality-invariant representations by utilizing relationship between paired samples and making the distributions similar. (B) We obtain decision boundaries from observed modal samples. (C) We can classify unseen modal samples by the learned boundaries through modality invariant representations.}
     \label{fig:concept}
  \vspace{-3mm}
\end{figure}
\vspace{-3mm}
\section{Introduction}
Learning modality-invariant representation is highly important in computer vision. Modality-invariant representations are interchangeable with different modalities in a classification problem. We aim to learn modality-invariant representation from two modalities $x \in X$ and $y \in Y$. Given paired samples with label $ {\bigl\{({x_i},{y_i},{t_i})\bigr\}^{n}_{i=1}\in \mathbb{R}^{d_x}\times\mathbb{R}^{d_y}\times\mathcal{C}}$, our goal is to learn the mapping to a common space, $G_x: \mathbb{R}^{d_x} \rightarrow\mathbb{R}^{d_{z}}$ , $G_y: \mathbb{R}^{d_y} \rightarrow\mathbb{R}^{d_{z}}$ and a classifier $h: \mathbb{R}^{d_z} \rightarrow\mathcal{C}$ such that both classifier $h\circ G_{x}$ and $h\circ G_{y}$ work well. For example, with regard to Mir Flickr dataset \cite{huiskes2008mir}, given paired samples of image features, tag features and corresponding labels, we want to learn the representations which satisfy the condition above.  We define such representations $f_x = G_x(x)\in\mathbb{R}^{d_z}$ and $ f_y=G_y(y)\in \mathbb{R}^{d_z}$ as modality-invariant representations. 

Previous works \cite{srivastava2012multimodal,srivastava2012learning,ngiam2011multimodal} had been done mainly to learn rich representation, provided complete sets of samples $ \bigl\{({x_i},{y_i},{t_i})\bigr\}^{n}_{i=1}$. However, collecting a large number of complete sets involves high labor cost. On the other hand, it is relatively easy to collect incomplete sets of samples such as $\bigl\{({x_i},{y_i})\bigr\}$, $\bigl\{({x_i},{t_i})\bigr\}$ and $\bigl\{({y_i},{t_i})\bigr\}$. We can handle such incomplete sets of samples appropriately using the modality-invariant representation. For example, we first learn modality-invariant representation $f_x$ and $f_y$ from a set of paired samples $ {\bigl\{({x_i},{y_i})\bigr\}^{n}_{i=1}}$. Then, we train a classifier on one modality and their labels $ {\bigl\{(f_{x_{j}} ,{t_j})\bigr\}^{m}_{j=1}}$. Finally, we can apply the classifier to the testing of $f_y$. Note that we can utilize the complete sets of samples with our proposed method. We show the illustration of our method in Fig. \ref{fig:concept}.

The important properties of modality-invariant representations are the following two things.
\begin{enumerate}
\setlength{\parskip}{0cm}
  \setlength{\itemsep}{0cm} 
\item Modality-paired samples are placed close with each other in the representations.
\item Two distributions of the representations are made similar.
\end{enumerate}
Learning the representation satisfying the two properties is important for the following two reasons. First, we can obtain discriminative features and place paired samples close by utilizing the relationship between paired samples. Second, obtaining similar distribution should improve the performance of a classifier when we train the classifier on one modality and test on another. This is because decision boundaries are considered to be drawn between the distribution of one class and another. Therefore, we can make good decision boundaries for appropriately classifying unseen modality samples if the distributions are similar.

There exist a large number of work on multimodal learning methods \cite{srivastava2012multimodal,srivastava2012learning, ngiam2011multimodal}, which aim to learn fusing multiple modal features by utilizing the relationship between paired samples. These researches focus on obtaining rich features from multiple modalities and do not consider divergence between distributions.
Also, Canonical Correlation Analysis (CCA) is an effective method of learning the relationship between paired samples. CCA projects different modality samples into a common space where the correlation between paired samples is maximized. However, CCA do not consider the divergence between two distributions. 

On the other hand, in Domain Adaptation (DA), many methods are proposed to reduce the divergence between source samples' distribution and target samples' distribution \cite{ben2010theory,long2015learning,ganin2014unsupervised}. They aim to construct domain-invariant classifier by making the distribution of different domains similar. However, to the best of our knowledge, existing DA method cannot deal with different modalities. Thus, within the framework of DA, we cannot deal with the relationship between modality-paired samples.

In this paper, we propose a method to learn the modality-invariant representations, which satisfy the two important properties above. We incorporate the idea of matching paired samples and DA to obtain the representations. We achieve it by  adversarial training, that is, simultaneously optimizing two models: a modality-discriminator ($D$) and a generator ($G$). $G$ aims to learn modality-invariant representation. $D$ estimates the probability that a sample came from each modality. The training procedure for $G$ is to maximize the probability of $D$ making a mistake and simultaneously minimize the distance between paired samples. We call our proposed method as Deep Modality Invariant Adversarial Network (DeMIAN), which aims to obtain a modality-invariant representation through adversarial training.
Our contributions are as follows,
\begin{itemize}
\setlength{\parskip}{0cm}
  \setlength{\itemsep}{0cm} 
\item We proposed a novel algorithm for modality-invariant representation learning, named Deep Modality Invariant Adversarial Network (DeMIAN).
\item Using the modality-invariant representations learned by DeMIAN, we achieved better classfication accuracy than with the state-of-the-art methods.
\item In the experiment of zero-shot learning, our proposed model achieved state-of-the-art accuracy for datasets that are considered as the benchmark datasets of zero-shot learning.
\end{itemize}
\vspace{-5mm}
\section{Related Works}
\subsection{Multimodal Learning}
In previous works on multimodal learning, deep Boltzmann machine (DBM), deep belief net and autoencoder were used because these non-linear generative models can extract a unified representation that fuses modalities together \cite{srivastava2012multimodal,srivastava2012learning,ngiam2011multimodal}. They aim to extract rich information from multiple modal samples by using the relationship between paired samples. Ngiam \etal \cite{ngiam2011multimodal} used CCA to learn the latent space between audio and video features, and trained a classifier by using only one modality and tested it on the other modality. They defined this problem as Shared Representation Learning (SRL). This problem situation is the same as ours (hereinafter, referred to as  SRL). According to \cite{chu2013sparse}, the formulation of CCA can be viewed as the minimization of distance between paired modalities in a latent space on the condition that their norm is equal to one. That is, 
\begin{equation}
\setlength{\abovedisplayskip}{2.5pt} 
\setlength{\belowdisplayskip}{2.5pt} 
\min_{w_x,w_y} \|X^{T}w_{x} - Y^{T}w_{y}\|^2_2
\end{equation}
\begin{equation}
\text{subject to} \;\;{w_{x}}^TXX^T{w_{x}} = 1,{w_{y}}^TYY^T{w_{y}} = 1
\end{equation}
In our algorithm, we propose to utilize the relationship between paired samples by minimizing the distance as used in the formulation of CCA. Moreover, we add the term which makes the distribution of different modalities similar. Thus, our model efficiently incorporates a modality-invariant factor with multimodal learning.
\vspace{-1mm}
\subsection{Domain Adaptation}
In DA, we aim to learn from an abundant labeled source data distribution and build a well-performing model on a different target data distribution. 
As we do not have any labeled sample for the target domain during the training, we have to employ a mechanism that extracts the domain-invariant feature. 
For this purpose, David \etal\cite{ben2010theory} proposed to reduce the divergence between the distribution of the source and the target space. A large number of methods using this idea have been proposed to extract the domain-invariant representation. Long \etal\cite{ajakan2014domain} proposed a CNN architecture for DA, where they introduced minimizing Maximum Mean Discrepancy (MMD) between the source and the target domains. 
Ganin \etal\cite{ganin2014unsupervised} introduced the idea of Generative Adversarial Network \cite{goodfellow2014generative} for DA. They used adversarial training for the domain-invariant feature extraction, which distinguishes which domain middle features are generated. As their method is closely related to our method, we describe their method in detail.
They made two branches from one layer of the CNN: one is the source samples' classifier network which includes a feature extractor Network, and the other is domain-classifier Network, which distinguishes the features' domain. To classify the target samples, they trained their network to simultaneously classify the source samples and to deceive the domain-classifier. They defined loss the function as 
\begin{equation}
E({\theta}_f,{\theta}_y,{\theta}_d) = L({\theta}_f,{\theta}_y)- \lambda L_{d}({\theta}_f,{\theta}_d)
\end{equation}
where ${\theta}_f$ is the parameter of the feature extractor that generates the middle feature, ${\theta}_y$ is the parameter of the classifier of the CNN,. ${\theta}_d$ is the parameter of the domain-classifier, $L({\theta}_f,{\theta}_y)$ is the loss for the source domain label, and $L_d({\theta}_f,{\theta}_d)$ is loss for the domain label. During the training, domain-classifier learns to maximize $E({\theta}_f,{\theta}_y,{\theta}_d)$ and the feature extractor learns to minimize it. Through this adversarial training, one can observe that the distribution of different domains matches in their middle feature space.

Ganin \etal \cite{ganin2014unsupervised} discussed the relationship between their algorithm and the ${\cal H} {\mathrm {\bigtriangleup}} {\cal H}$-distance, which is widely used in the theory of non-conservative DA \cite{ben2010theory}. The ${\cal H} {\mathrm {\bigtriangleup}} {\cal H}$-distance defines a discrepancy between two distributions ${\cal S}$ and ${\cal T}$ w.r.t. a hypothesis set ${\cal H}$. Using this distance, we can obtain a probabilistic bound on the performance of some classifier on the target domain. Ganin \etal showed an upper bound for their model in their equation (13) \cite{ganin2014unsupervised}. This theorem cannot be directly applied to modality-invariant representation learning because this theorem is for the same modality. The prerequisite for utilizing  the ${\cal H} {\mathrm {\bigtriangleup}} {\cal H}$ -distance for bounding error on the target domain is the existence of an ideal joint hypothesis \cite{ben2010theory}. That is, if there exists a hypothesis that minimizes the combined error of the source and target domains well, we can measure the adaptability of a source-trained classifier by using the H-divergence between the marginal distributions $D_{\cal S}$ and $D_{\cal T}$. This hypothesis is usually thought to be satisfied for datasets used in DA. When extracting discriminative information, domain-shift is considered to be moderate despite the difference in style or context. However, regarding multimodal learning, we cannot make such an extreme hypothesis because of modality-specific properties. Furthermore, owing to the difference in dimension between modalities, we have to find some latent space with a different network. Our model solves this difficulty by finding a latent space where the distance of paired samples' is minimized. We assume that by minimizing paired samples' distance, the prerequisite for the ${\cal H} {\mathrm {\bigtriangleup}} {\cal H}$-distance is satisfied in our algorithm. 
\vspace{-1mm}
\subsection{Zero-Shot learning}
Zero-shot learning deals with the problem of learning to classify previously unseen class instances. This task is highly important because even in large-scale classifications, the labels for many instances or categories can often be missing. We consider a version of the zero-shot learning problem where the seen class source and target domain data are provided. The goal  is to accurately predict the class label of an unseen target domain instance on the basis of the revealed source domain side information (e.g. attributes) for unseen classes. Previous methods focus on learning the latent embeddings of the target and the source domain based on similarity or other metrics \cite{bucher2016improving,zhang2016zeroeccv,zhang2016zero,zhang2015zero}, and some work deal with this problem using the concept of DA \cite{kodirov2015unsupervised}. 
To learn the modality-invariant representation, we propose a novel method that integrates multimodal learning and DA. We leverage the relationship between different modalities to utilize the idea of DA for multimodal learning. Our method is shown to be effective for SRL problem and zero-shot learning. 
\begin{figure}[t]
  \begin{center}
   \includegraphics[width=0.9\hsize,height=0.7\hsize]{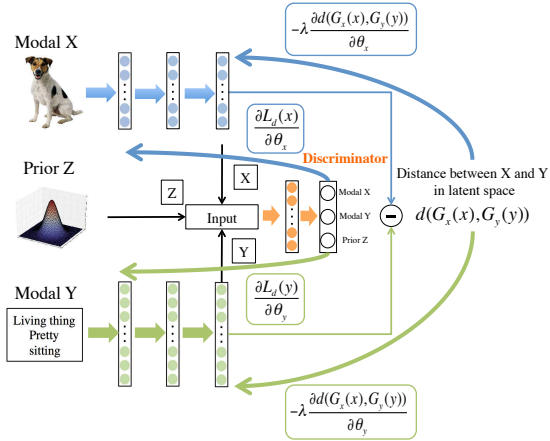}
  \end{center}
    \caption{The proposed model: Deep Modality Invariant Adversarial Network. We aim to learn modality-invariant representations by using a minimax two-player game involving the generator and the discriminator.}
  \label{fig:propose}
\end{figure}
\vspace{-3mm}
\section{Proposed Method}
\vspace{-1mm}
In this section, we now describe the details of the proposed model. We first explain about the problem setting and requirements for our model, and then we explain about the components of our model for satisfying the requirements. Finally, we give an explanation about the learning procedure. We describe the overview of our model in Fig. \ref{fig:propose}.

Given samples $ {\bigl\{({x_i},{y_i},{t_i})\bigr\}^{n}_{i=1}\in \mathbb{R}^{d_x}\times\mathbb{R}^{d_y}\times\mathcal{C}}$, our goal is to learn mapping to common space, $G_x: \mathbb{R}^{d_x} \rightarrow\mathbb{R}^{d_{z}}$ , $G_y: \mathbb{R}^{d_y} \rightarrow\mathbb{R}^{d_{z}}$ and a classifier $h: \mathbb{R}^{d_z} \rightarrow\mathcal{C}$ such that both classifier $h\circ G_{x}$ and $h\circ G_{y}$ work well. We first learn mapping using paired samples  ${\bigl\{({x_i},{y_i})\bigr\}^{n}_{i=1}}$, then train a classifier by learned representations. 
The following two things are required for obtaining modality-invariant representation,
\vspace{-2mm}
\begin{enumerate}
\setlength{\parskip}{0cm}
  \setlength{\itemsep}{0cm} 
\item We have to project samples into common space by considering relationship between paired samples.
\item We have to make the distribution of ${f_x}, {f_y}$ similar. (We utilize the idea of DA)
\end{enumerate}
As for the first requirement, we aim to satisfy the assumption of DA and to obtain discriminative information by making use of the relationship between paired samples. With regard to the second requirement, from the perspective of DA, we assume that we can train a classifier which works well for both modalities by making the distribution similar. We have to construct the model which simultaneously satisfies the two requirements.  

Our proposal is for learning representations from  ${\bigl\{({x_i},{y_i})\bigr\}^{n}_{i=1}}$. We call the function $G_x, G_y$ as generators.
We denote the parameters of generators as $\theta_x, \theta_y$. i.e. ${f_x} = G_x(x;\theta_x)$,${f_y}=G_y(y;\theta_y)$.
\vspace{-1mm}
\subsection{Generator to consider relationship between modalities}
For the first requirement, we define the objective for this matching as 
\begin{equation}
\setlength{\abovedisplayskip}{2.5pt} 
\setlength{\belowdisplayskip}{2.5pt} 
J(\theta_x,\theta_y) = \sum_{i=1}^n d({f_x}, {f_y})
\end{equation}
where we used Euclidean distance or cosine distance for $d({f_x}, {f_y})$, which can consider the matching of paired modality as used in CCA. Here, we define the distribution of  ${f_x}$ and ${f_y}$ as ${\cal S}({f_x}) = {\bigl\{{G_x(x;\theta_x) | x\sim {\cal S}(x)}\bigr\}}$ and ${\cal T}({f_y}) = {\bigl\{{G_y(y;\theta_y) | y\sim {\cal T}(y)}\bigr\}}$ respectively.
\subsection{Modality-discriminator to make the distribution similar}
As for the second one, we have to measure the dissimilarity of distributions ${\cal S}({f_x})$ and ${\cal T}({f_y})$. However, measuring the dissimilarity is non-trivial, given that ${f_x}$ and ${f_y}$ is high-dimensional, and that distributions of themselves are changing constantly as learning progresses. Hence, we utilize the modality-discriminator, $D_d$ with the parameters $\theta_d$. We can estimate the dissimilarity by looking at the loss of $D_d$, provided that $D_d$ has been trained to discriminate between ${f_x}$ and ${f_y}$ well. 

Therefore, we seek the parameters $\theta_x$ and $\theta_y$ that maximize the loss of $D_d$, while simultaneously seeking the parameters $\theta_d$ that minimize the loss of $D_d$. This is the adversarial training method for our model. Actually, we seek to minimize the loss of $J(\theta_x,\theta_y)$ in addition to the loss for adversarial training.

To effectively utilize adversarial training, we propose to input gaussian samples $ z\sim N(0,I_{z_d})$ to $D_d$ as well as ${f_x}$ and ${f_y}$. Given the setting of minimax problem between generators and a discriminator, the difficulty is that generators can easily achieve a draw. For example, generators can always achieve a draw if it generates ${f_x}$ and ${f_y}$ with all elements zeros. Actually, as the objective contains the term concerning relationship between paired samples, such an extreme case will not happen. This insight, however, indicates the fact that generator can return non-discriminative representations to deceive a discriminator. To avoid it, we consider $ z\sim N(0,I_{z_d})$ and ${f_x}$ and ${f_y}$ as being generated from different modality then solve the minimax problem. Since we always sample random gaussian value $ z\sim N(0,I_{z_d})$, ${f_x}$ and ${f_y}$ should be placed isotropically within gaussian distribution to deceive a discriminator. We will confirm the effect of $z$ in the experiment section. 
\begin{algorithm}[t!]                    
\caption{Optimization of DeMIAN}         
\label{alg1}                          
\begin{algorithmic}                  
\FOR{number of iterations}
\FOR{k steps}
\STATE $\bullet$ Sample mini-batch of $m$ Gaussian samples $\left\{{z_{1},....z_{m}}\right\}$\\
\STATE $\bullet$ Sample mini-batch of $m$ paired samples from real data distribution $\left\{x_{1},....x_{m}\right\}$, $\left\{y_{1},....y_{m}\right\}$\\
\STATE $\bullet$ Update the discriminator by ascending its stochastic gradient
\begin{equation}
\begin{split}
\nabla_{\theta_{d}}\bigl[&\log {(1-D_d(G_x{(x_{i})}=1))}\\
	+&\log {(1-D_d(G_y{(y_{i})}=2))}\\
	+&\log {(1-D_d(z_{i}=3))}\bigr]
\end{split}
\end{equation}
\ENDFOR
\STATE $\bullet$ Update the generator by descending its stochastic gradient
\begin{equation}
\begin{split}
\nabla_{\theta_{g}}\bigl[&\log {(1-D_d(G_x{(x_{i})}=1))}\\
	+&\log {(1-D_d(G_y{(y_{i})}=2))}\\
	+&\log {(1-D_d(z_{i}=3))}\\
	+&\lambda d(G_x{({x_{i}})},G_y{({y_{i}})})\bigr]
\end{split}
\end{equation}
\ENDFOR
\end{algorithmic}
\end{algorithm}
\begin{figure*}[t!]
\centering
   \begin{subfigure}[CCA]{
   \centering
   \includegraphics*[width=0.23\hsize]{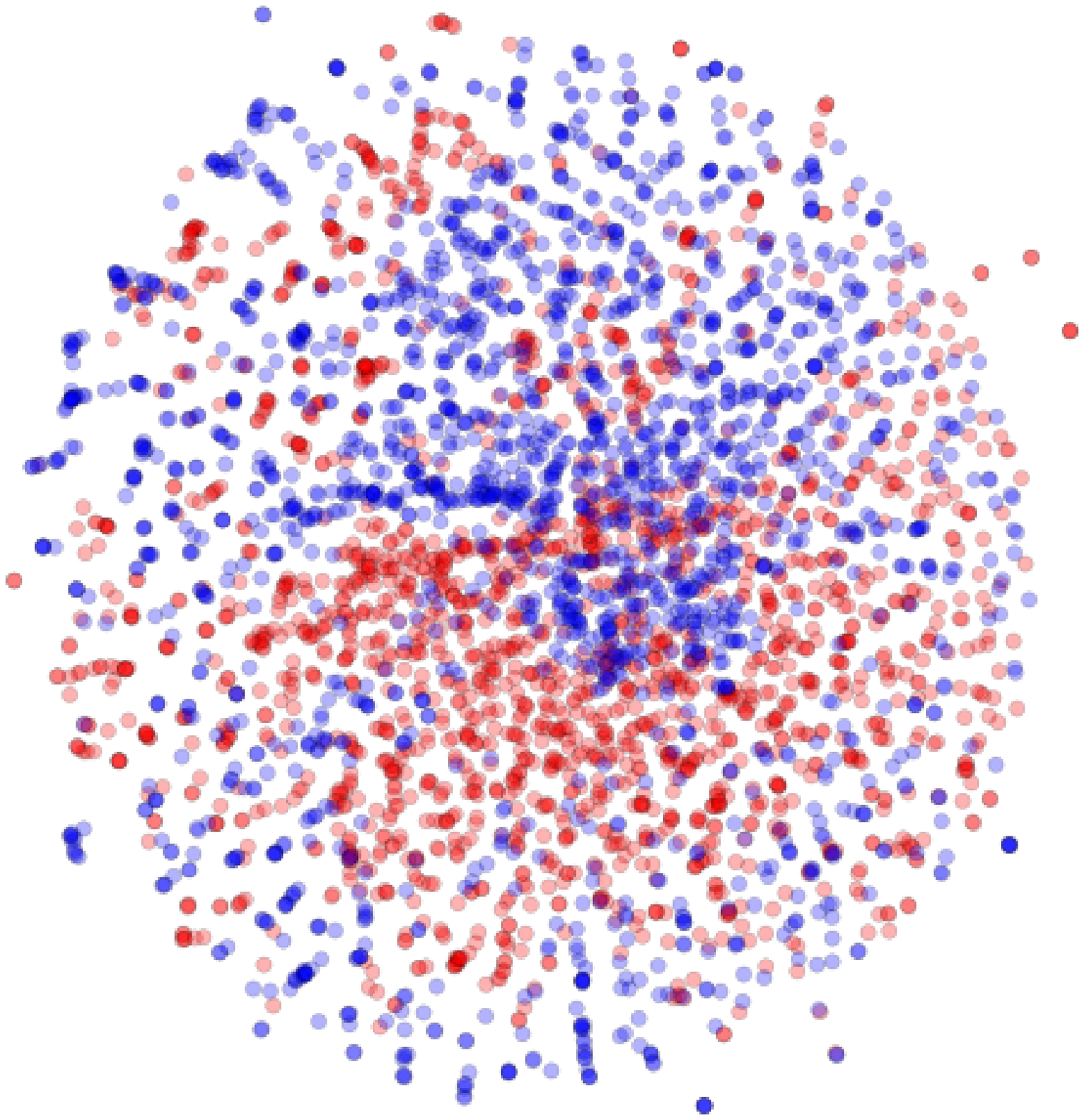} \label{fig:mnist_cca}
}
    \end{subfigure}
 \centering
\begin{subfigure}[MIAN]{
 \centering
   \includegraphics*[width=0.23\hsize]{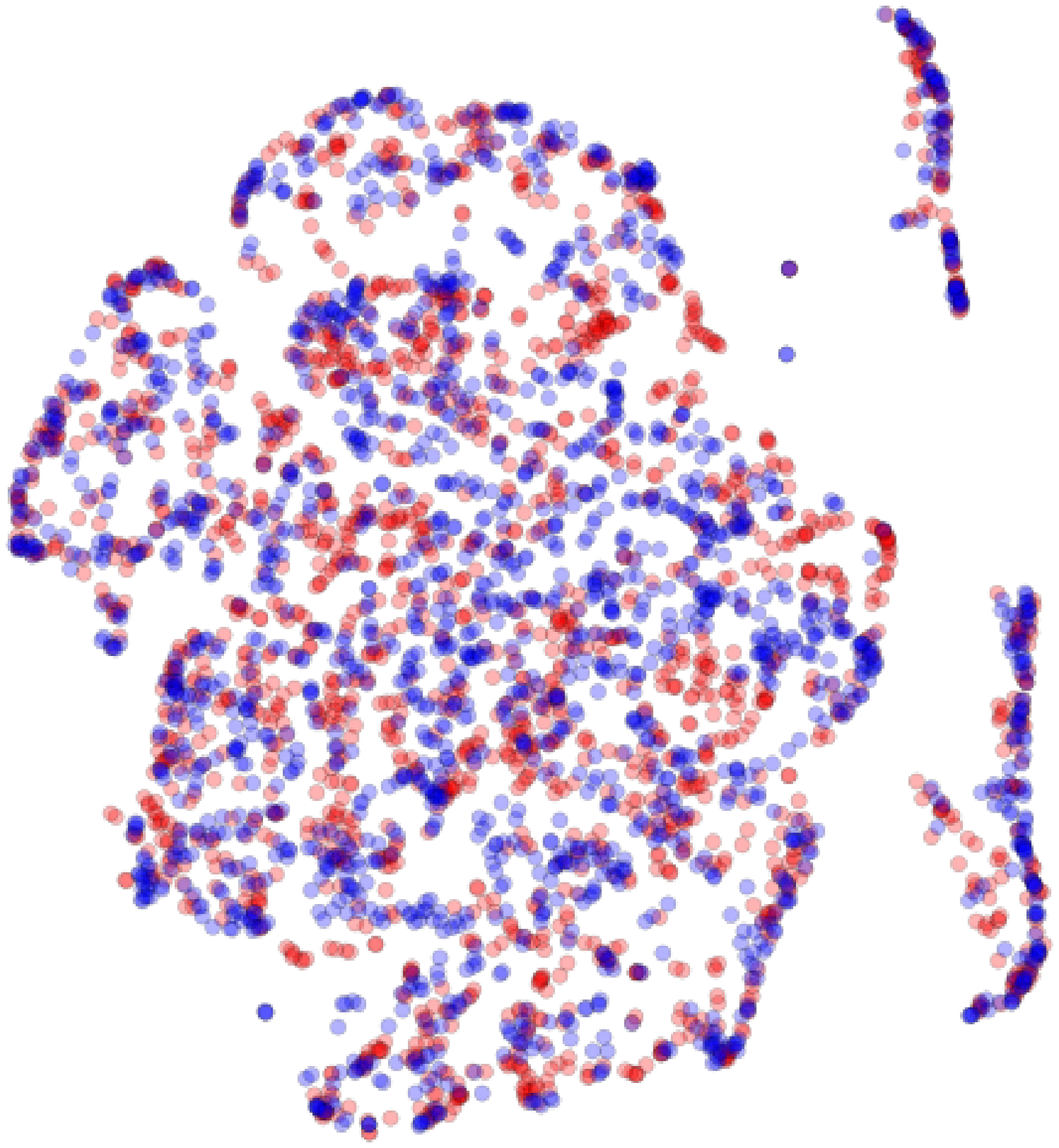}  \label{fig:mann_shallow}}
    \end{subfigure}
  \centering
   \begin{subfigure}[DeMIAN]{
    \centering
   \includegraphics*[width=0.23\hsize]{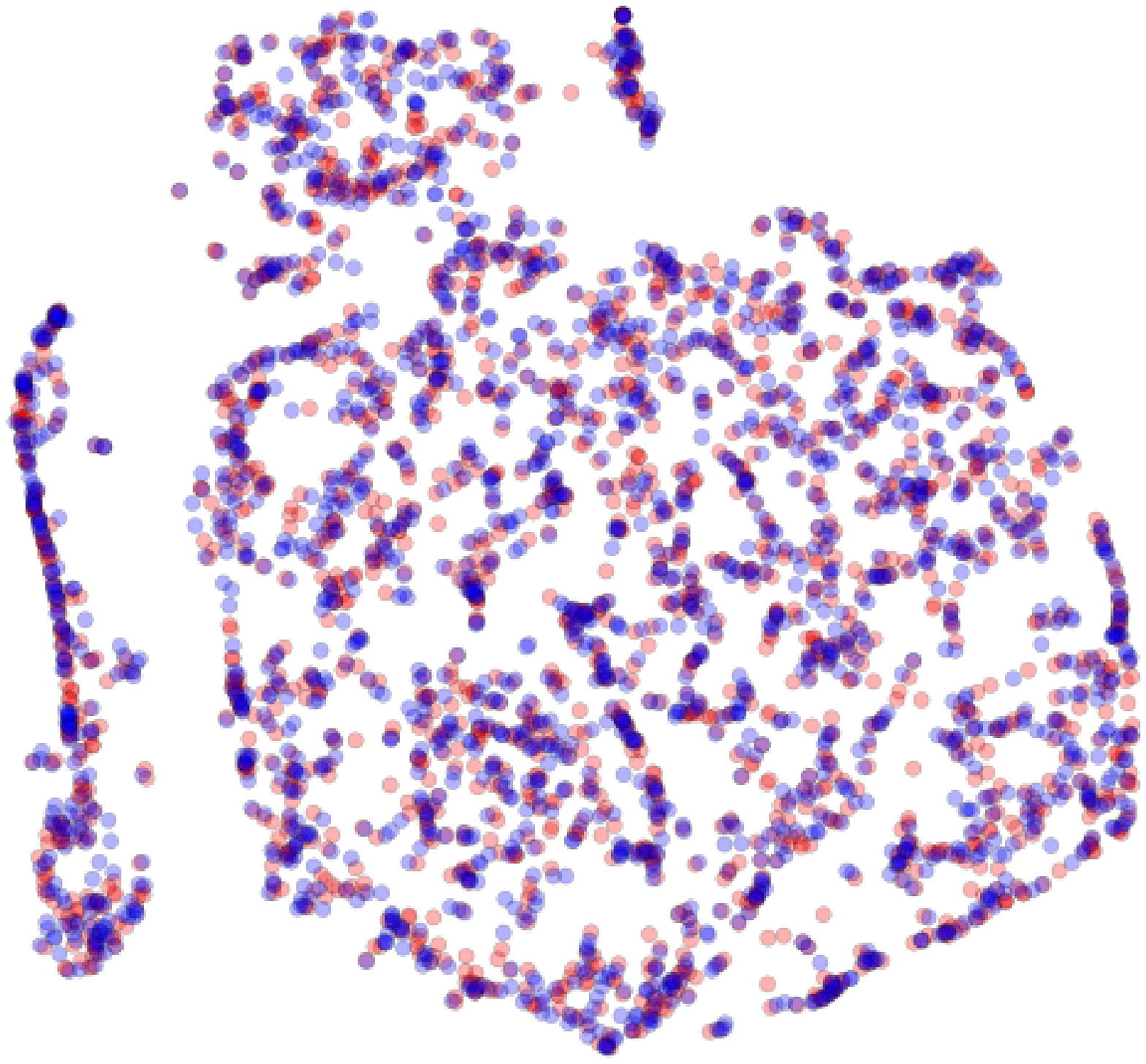}
     \label{fig:mian_mnist_deep}}
    \end{subfigure}
    \begin{subfigure}[DeMIAN without prior $z$]{
    \centering
   \includegraphics*[width=0.23\hsize]{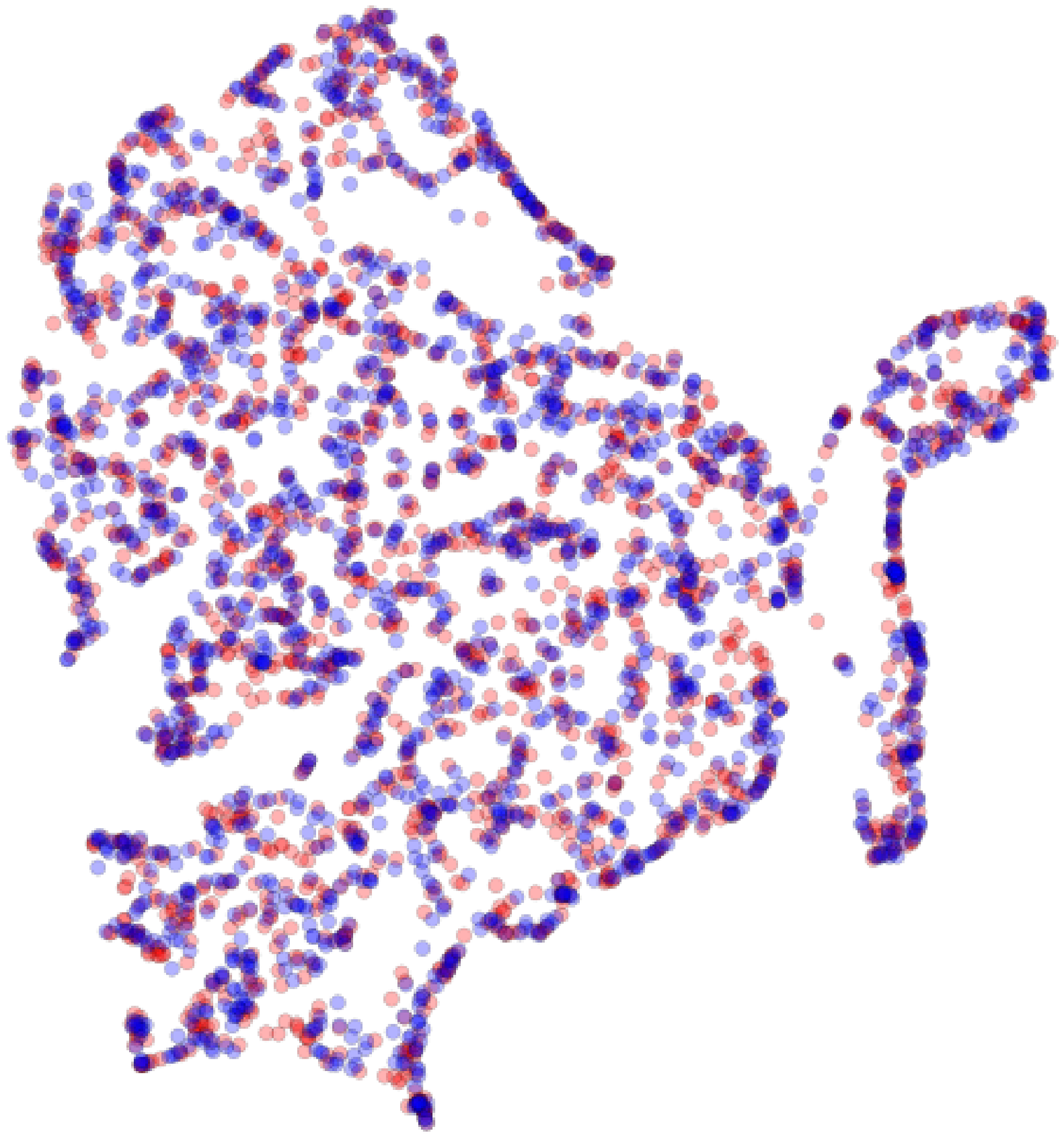}
    \label{fig:mian_noz}
}
    \end{subfigure}
  \caption{Comparison of the embedding in the MNIST experiments.}
 \label{fig:embed_mnist}
\end{figure*}
\begin{figure}[t!]
 \centering
   \begin{subfigure}[Training with {\bf left} samples]{
    \centering
   \includegraphics*[width=0.43\hsize]{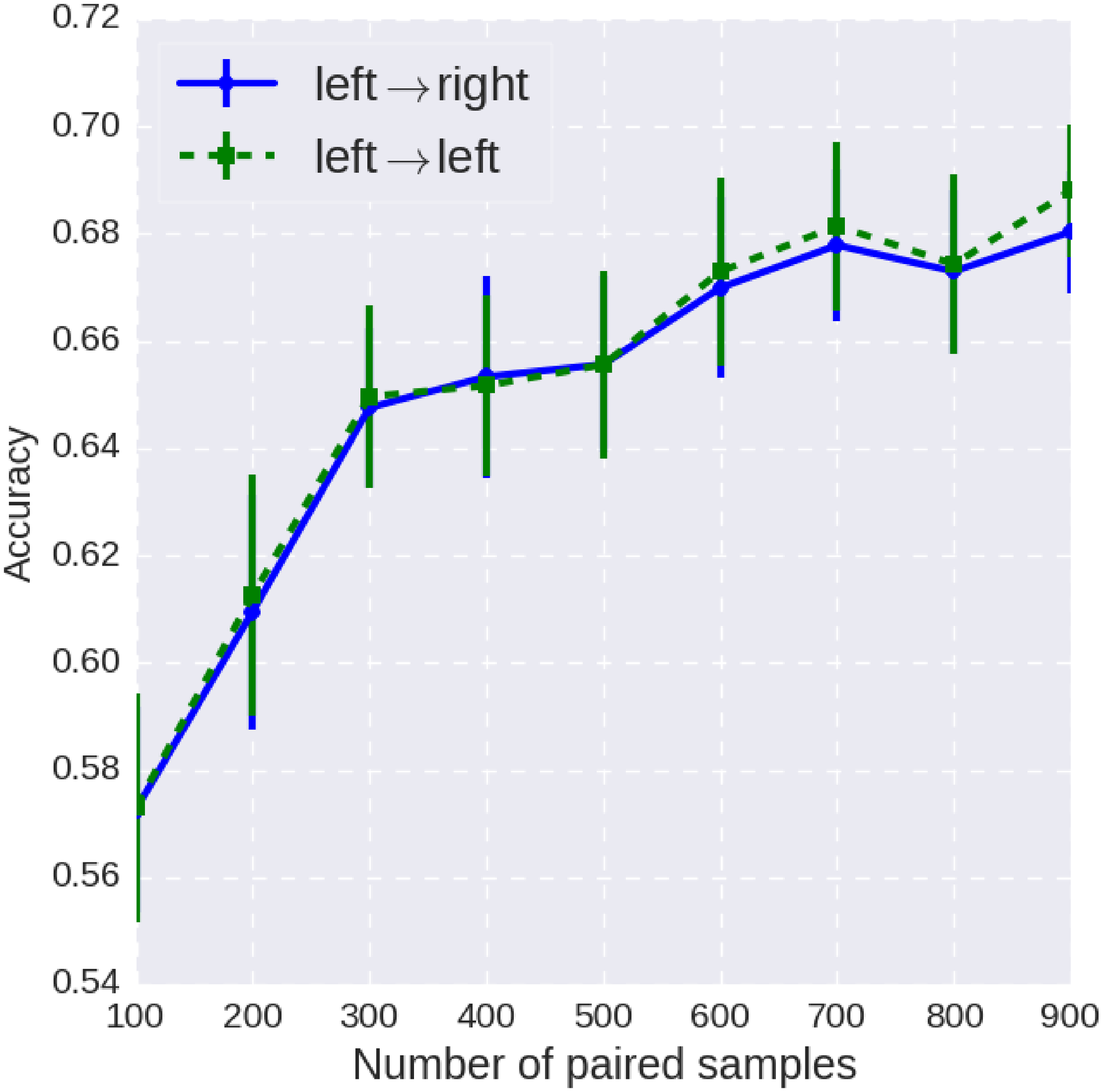}}
   \end{subfigure}
   \begin{subfigure}[Training with {\bf right} samples]{
    \centering
   \includegraphics*[width=0.43\hsize]{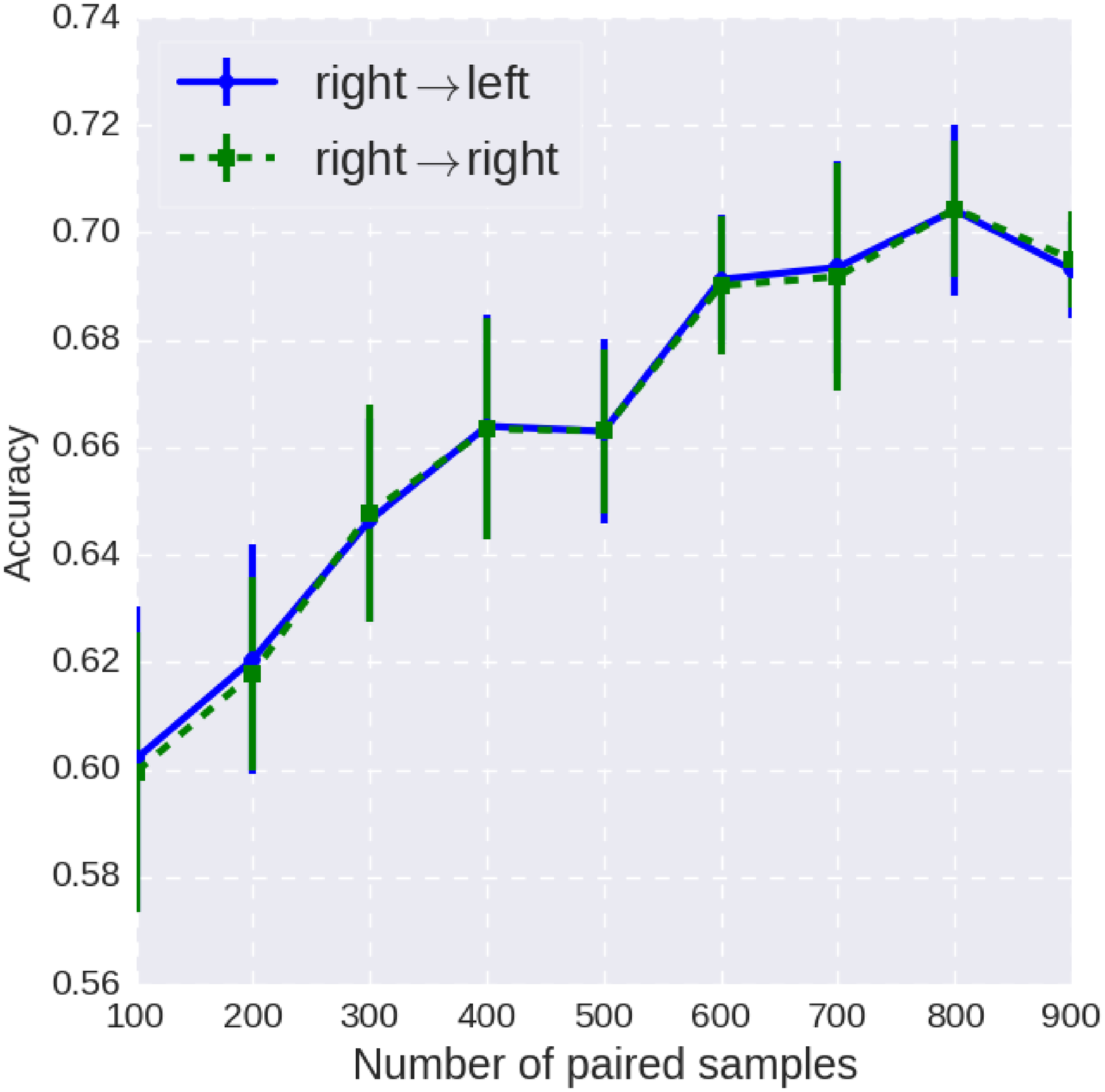}}
    \end{subfigure}
  \caption{Relationship between the number of training samples and accuracy}
 \label{fig:graph_mnist}
\end{figure}
\vspace{-1mm}
\subsection{Formulation of DeMIAN and its optimization}
\vspace{-1mm}
Given the discussion above, we show the objective for our model as follows, 
\begin{equation}
\begin{split}
 z_i&\sim N(0,I_{z_d})\\
L(D_d({x_{i}},{y_{i}}),t_m) =& \log {(1-D_d(G_x{(x_{i})}=1))}\\
	+&\log {(1-D_d(G_y{(y_{i})}=2))}\\
	+&\log {(1-D_d(z_{i}=3))}\\
E_d(\theta_d) = &-L(D_d({x_{i}},{y_{i}}),t_m)\\
E_g(\theta_x,\theta_y) =&J(\theta_x,\theta_y) - \lambda L(D_d({x_{i}},{y_{i}}),t_m)
\end{split}
\end{equation}
where $t_m$ means the modality label. $\lambda$ is a parameter for balancing the loss of multimodal matching and adversarial training. In our work, we decide this parameter by validation split of a dataset. 
Under these objective functions, we seek parameters that satisfy,
\begin{eqnarray}
\setlength{\abovedisplayskip}{0pt}
\setlength{\belowdisplayskip}{0pt} 
\hat{\theta}_x,\hat{\theta}_y &=& \argmin_{\theta_{x},\theta_{y}}  {E_g(\theta_{x},\theta_{y},\theta_{d})}\\
\hat{\theta}_d &=& \argmax_{\theta_d} E_d(\theta_d)
\vspace{-3mm}
\end{eqnarray}
\begin{table}[t!]
\vspace{-1mm}
\centering
  \begin{tabular}{|c|c|c|c|c|c|} \hline
      &CCA& KCCA& DCCA& MIAN &DeMIAN\\ \hline
      Dev & 28.1 & 33.5&--&43.8&48.1 \\\hline
    Test & 28.0 & 33.0 &47.0&43.7&{\bf 48.0}\\ \hline
  \end{tabular}
   \vspace{1mm}
  \caption{Correlation value of top50 dimensions'  between the features from the left and the right digits of MNIST. The values of CCA, KCCA were from \cite{andrew2013deep}, whereas that of  DCCA was from \cite{wang2015stochastic}. We used non-linear model of MIAN.}
   \label{tb:value_mnist}
   \vspace{-2.mm}
\end{table}
At the saddle point, the parameters $\theta_{x},\theta_{y}$ minimize the modality classification loss while minimizing the loss for matching paired samples.

As the activation function, we used ReLU or ELU \cite{clevert2015fast} and used BatchNormalization (BN) \cite{ioffe2015batch} after the activation. BN is known to be  highly effective for optimizing Generative Adversarial Nets \cite{radford2015unsupervised}, and we confirmed that BN can also stabilize and improve the performance of our model. As BN can properly control the scale of output in each layer, it should be effective for matching  the distribution of ${f_x}, {f_y}$ and $z$. For the discriminator, we used ReLU for activation in all the experiments.
In Algorithm \ref{alg1}, we show the optimizing procedure of our method. We set $k=1$ in all of our experiments, but we think that this value should be changed according to the datasets or tasks. 
\begin{table}[t!]
\centering
  \begin{tabular}{|c|c|c|c|c} \hline
& \multicolumn{2}{|c|}{ {\bf \ Training modality}$\rightarrow$}\\
& \multicolumn{2}{|c|}{ {\bf Testing modality}}\\\hline
      Method &Left $\rightarrow$ Right&Right $\rightarrow$ Left\\ \hline
      CCA & 0.703 & 0.675 \\\hline
    MIAN (non-linear) &0.754&0.713\\\hline
      DeMIAN w/o {\it z} &0.680&0.761\\\hline
    DeMIAN &{\bf 0.810}&{\bf 0.795}\\\hline
  \end{tabular}
   \vspace{1mm}
  \caption{Result of the Mnist Recognition experiment. Training modality means the input modality for the supervised training, whereas Testing modality means the input modality when testing. }
  \label{tb:reg_mnist}
\end{table}
\begin{table*}[h!]
\centering
  \begin{tabular}{|c|c|c||c|c|c|c} \hline
&  \multicolumn{5}{|c|}{ {\bf Training modality} $\rightarrow${\bf Testing modality}}\\\hline
     Method&Tag $\rightarrow$ Image&Image $\rightarrow$ Tag& Tag $\rightarrow$Tag&Image $\rightarrow$Image &Tag and Image$\rightarrow$Tag and Image\\ \hline 
      DBM \cite{srivastava2012multimodal} &--&--&--&--&0.528\\\hline
      CCA & 0.312 & 0.404&0.428&0.381&0.496 \\\hline
    Deep CCA & 0.438 &0.455&0.463&0.464&0.570\\ \hline
       MIAN (linear) &0.458&0.438& {\bf 0.528}&0.548&0.598\\\hline
    DeMIAN &{\bf 0.544}&{\bf 0.487}&0.512&{\bf 0.567}&{\bf 0.599}\\\hline
  \end{tabular}
   \vspace{1mm}
  \caption{Result of Mir Flickr recognition experiment based on MAP . {\bf Training modality} means the input modality when training a linear classifier. {\bf Testing modality} means the input for the testing. Note that we show the result where we had access to 25,000 labeled samples.}
  \label{tb:mirflickr}
\end{table*}
\begin{table*}[t!]
\centering
  \begin{tabular}{|c|c|c||c|c|c|c} \hline
 & \multicolumn{4}{|c|}{ {\bf Training modality} $\rightarrow${\bf Testing modality}}\\\hline
      Method&Image $\rightarrow$ Attribute& Attribute $\rightarrow$ Image&Image $\rightarrow$ Image&Attribute $\rightarrow$Attribute\\ \hline
      CCA & 0.100 &0.026&0.033&0.176 \\\hline
    MIAN (linear) &0.150&{\bf 0.046}&{\bf 0.082}& {\bf 0.186}\\\hline
    DeMIAN &{\bf 0.172}&0.044&  0.068&0.183\\\hline
  \end{tabular}
   \vspace{1mm}
  \caption{Result of the SUN Attribute recognition experiment.}
  \label{tb:reg_sun}
  \vspace{-3mm}
\end{table*}
\vspace{-2mm}
\section{Experiment}
\vspace{-2mm}
We tested our model by classification for SRL and zero-shot learning. For SRL, we used MNIST, SUN Attribute \cite{patterson2014sun} and Mir Flickr \cite{huiskes2008mir}. For zero-shot learning, we used SUN Attribute \cite{patterson2014sun} and Caltech-UCSD Birds-200-2011 (CUB-200-2011) \cite{wah2011caltech}, which are the benchmark image datasets for zero-shot learning. In all the experiments, we used Adam \cite{kingma2014adam} for optimization of our model, where we set $\beta_{1} = 0.5$ in all the experiments. Note that notation of DeMIAN in our result means our proposed model with 3 layers, that of MIAN means our model with 2 layers. MIAN includes the linear and non-linear models, which we will mention in an understandable way. We trained the logistic regression for the learned representations in SRL and trained multilayer-perceptron for zero-shot learning experiments.
\vspace{-2mm}
\subsection{MNIST}
\vspace{-2mm}
We divided the MNIST dataset into a left half and a right half as in \cite{andrew2013deep} to input our model separately. We used 60,000 samples for the training and 10,000 for the testing. We followed Andrew \etal, \cite{andrew2013deep}, wherein 6,000 samples of the training dataset were used as the validation split.  
We tested non-linear model of MIAN and DeMIAN in this experiment. We used ReLU as the activation function of the generator. The number of  units in our models was  [392, 300] for MIAN and [392, 1000, 50] for DeMIAN. As a function of $d(G_x{({x_{i}})},G_y{({y_{i}})})$, L2-squared distance was used, and we set $\lambda= 5.0$. For the discriminator, we used one hidden ReLU layer followed by a linear layer. We set learning rate of optimizer to $2.0\times10^{-4}$. A weight decay parameter of $1.0\times10^{-3}$ was used for all layers.  We set the mini-batch size to 500 during the training. We compared our proposed model with the model that does not input $z \sim N(0,I_z)$ to the discriminator. 

Table \ref{tb:value_mnist} shows the sum of the correlation value in the top 50 dimensions. The number of hidden units in MIAN was set to [392, 50].  Our model showed a significant improvement compared with the existing method. From this result, a matching distribution can effectively lead to a high correlation between multi-view features. Table \ref{tb:reg_mnist} shows the recognition experiment result learned by each model. Our proposed model achieved better performance compared with CCA. More interestingly, our model showed significantly better performance than our model without $z \sim N(0,I_z)$. It indicates that incorporating this prior can lead to better representation. Moreover, we visualized the learned representation using t-SNE \cite{maaten2008visualizing} in Fig. \ref{fig:embed_mnist}. From this figure, we can observe that the distribution between left half and right half digits matched densely compared with the embedding of CCA, and that the distribution of DeMIAN seemed to be more compact than that of MIAN. From Fig. \ref{fig:graph_mnist}, we can demonstrate the effectiveness of the latent space in providing modality-invariant features that allow for easier classification even when the number of labeled samples is limited. The performance when using only 1,000 samples was comparable to the performance when using all of the training samples. This indicates that our model performs well for datasets that do not have many labeled samples.
\vspace{-3mm}
\subsection{Mir Flickr}
\vspace{-2mm}
This data set consists of 1 million images from the social photography website Flickr along with their user-assigned tags. Twenty-five thousand images were annotated for 38 classes, where each image may belong to several classes \cite{huiskes2008mir}. We used 15,000 images for the training and 10,000 for the testing within labeled samples. Five thousand images of the training split were used for the validation. We used the mean average precision (MAP) for the evaluation following an existing work \cite{srivastava2012multimodal}.
Each tag input was represented using the vocabulary of the 2000 most frequently used tags. 
Images were represented by 3,857-dimensional features extracted by concatenating the pyramid histogram of words (PHOW) features \cite{bosch2007image}, Gist features \cite{oliva2001modeling}
and MPEG-7 descriptors \cite{manjunath2001color}(EHD, HTD, CSD, CLD, SCD). 

The number of units was [3857, 500] and [2000, 500] for the image feature and tag feature respectively for the linear model of MIAN , and  [3857, 1000, 200], [2000, 1000, 200] for DeMIAN. We used ReLU as the activation function of the generator. As a function of $d(G_x{({x_{i}})},G_y{({y_{i}})})$, L2-squared distance was used, and we set $\lambda= 5.0$ for both models. For the discriminator, we used one hidden ReLU layer with the same number of shared hidden units followed by one linear layer. We set learning rate of optimizer to $2.0\times10^{-3}$. A weight decay parameter of $1.0\times10^{-3}$ was used for all layers.  We set the mini-batch size to 500 during the training. For a comparison, we implemented Deep CCA \cite{andrew2013deep} in addition to CCA in this experiment. We used the optimization method proposed in \cite{wang2015stochastic} and used the same structure as our proposed method and added the BN layer for a fair comparison.

We show the result of Mir Flickr in Table \ref{tb:mirflickr}. In this table, we show the modality used for training and testing respecitively. Image $\rightarrow$ Tag means that we used image features and its labels for supervised training, and tested on Tag features. Image and Tag $\rightarrow$ Image and Tag means that we used both image features and Tag features for supervised training, and tested on both features. We averaged the output of the classifier for image features and tag features in this setting. Our model achieved better performance than other existing methods for both Image $\rightarrow$ Tag and Tag $\rightarrow$ Image. From the result of Image $\rightarrow$ Image and Tag $\rightarrow$ Tag, we can observe that our proposed adaptation did not worsen the performance on the source modality compared with CCA and DCCA\cite{wang2015stochastic}. Comparing CCA and MIAN (linear), we can see the direct effect of our modality adaptation method.
We can see that our model learned rich representations that are useful for both the source modality and the target modality. From the result of Image and Tag $\rightarrow$ Image and Tag, our model performed better than DBM \cite{srivastava2012multimodal}, which is one of the most successful models for multimodal learning. The result also showed that the representation from both modalities was effective for training a linear classifier. In this sense, our proposed model learned a modality-invariant rich representation.
\begin{table}[t]
\centering
  \begin{tabular}{|c|ll|} \hline
      Method &{CUB-200-2011}&{SUN Attribute}\\\hline
      Akata \etal\cite{akata2015evaluation} &{40.3}&{--} \\
      Kodirov \etal \cite{kodirov2015unsupervised}&{47.9}&{--} \\
      Peng \etal \cite{peng2016joint} &{49.87}&\\
      Lampert \etal\cite{lampert2014attribute} &{--}&{72.00}\\
      Paredes \etal \cite{romera2015embarrassingly} &{--}&{{$82.10\pm0.32$}}\\
      SSE-ReLU\cite{zhang2015zero} &{{$30.41\pm0.20$}}&{{$82.50\pm1.32$}}\\
      JLSE \cite{zhang2016zero} &{{$42.11\pm0.55$}}&{{$83.83\pm 0.29$}}\\
      Bucher \etal \cite{bucher2016improving} &{{$43.29\pm0.38$}}&{{$84.41\pm 0.71$}}\\
      Hard Negative \cite{bucher2016hard} &{{$45.87\pm0.34$}}&{{$86.21\pm 0.88$}}\\\hline
       DeMIAN &{{$\bf 57.5\pm0.56$}}&{{$\bf 87.6\pm1.3$}}\\\hline    
       \cite{romera2015embarrassingly} + SP-ZSR\cite{zhang2016zeroeccv} &--&{{$89.5$}}\\
       JLSE + SP-ZSR\cite{zhang2016zeroeccv} &{{$55.34\pm0.77$}}&{{$86.12\pm0.99$}}\\\hline
  \end{tabular}
  \vspace{1mm}
  \caption{Result of the zero-shot learning. Our model achieved state-of-the-art accuracy for the SUN and CUB-200-2011 dataset. Especially, SUN's score was the best one including the ensemble method.}
  \label{tb:sun_cub}
  \vspace{-3mm}
\end{table}
\begin{figure*}[h]
\centering
   \subfigure[VGG feature]{
 \centering
   \includegraphics*[width=0.3\hsize]{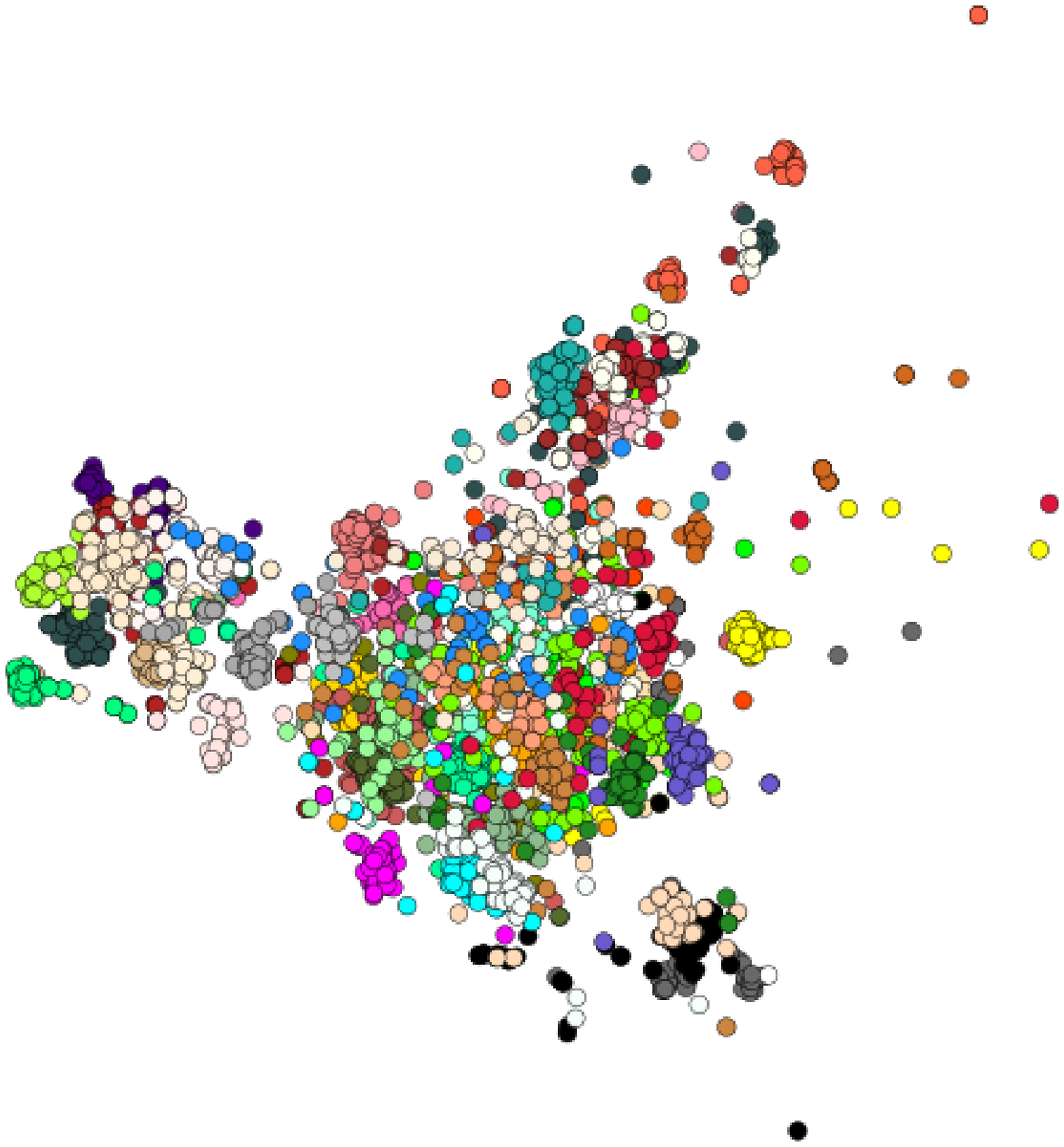}
   \label{fig:vgg_feature}}
 \centering
   \subfigure[DeMIAN representation]{
    \centering
   \includegraphics*[width=0.3\hsize]{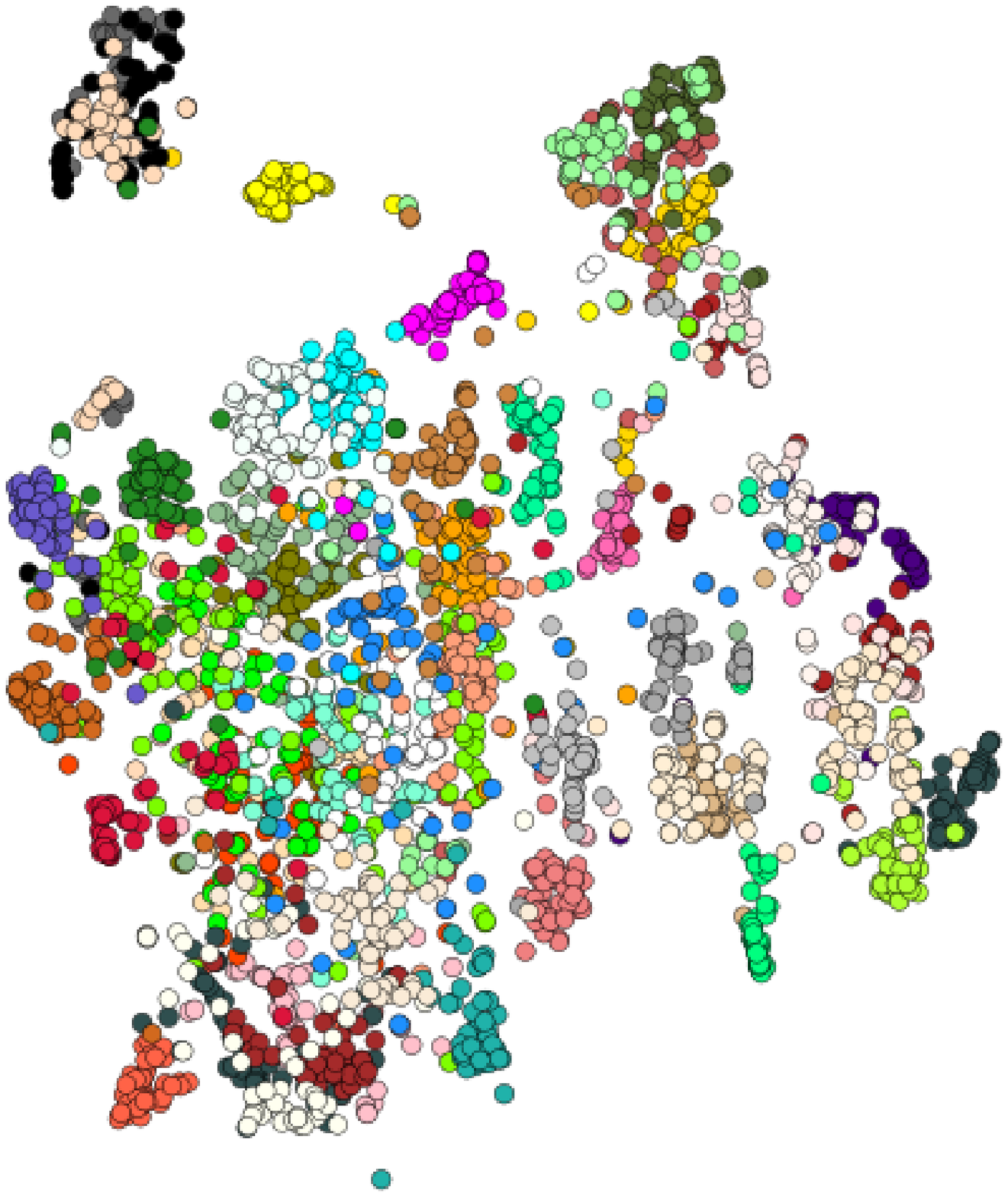}
   \label{fig:mian_cub}}
   \subfigure[Change in accuracy by the number of labeled attributes]{
    \centering
   \includegraphics*[width=0.3\hsize]{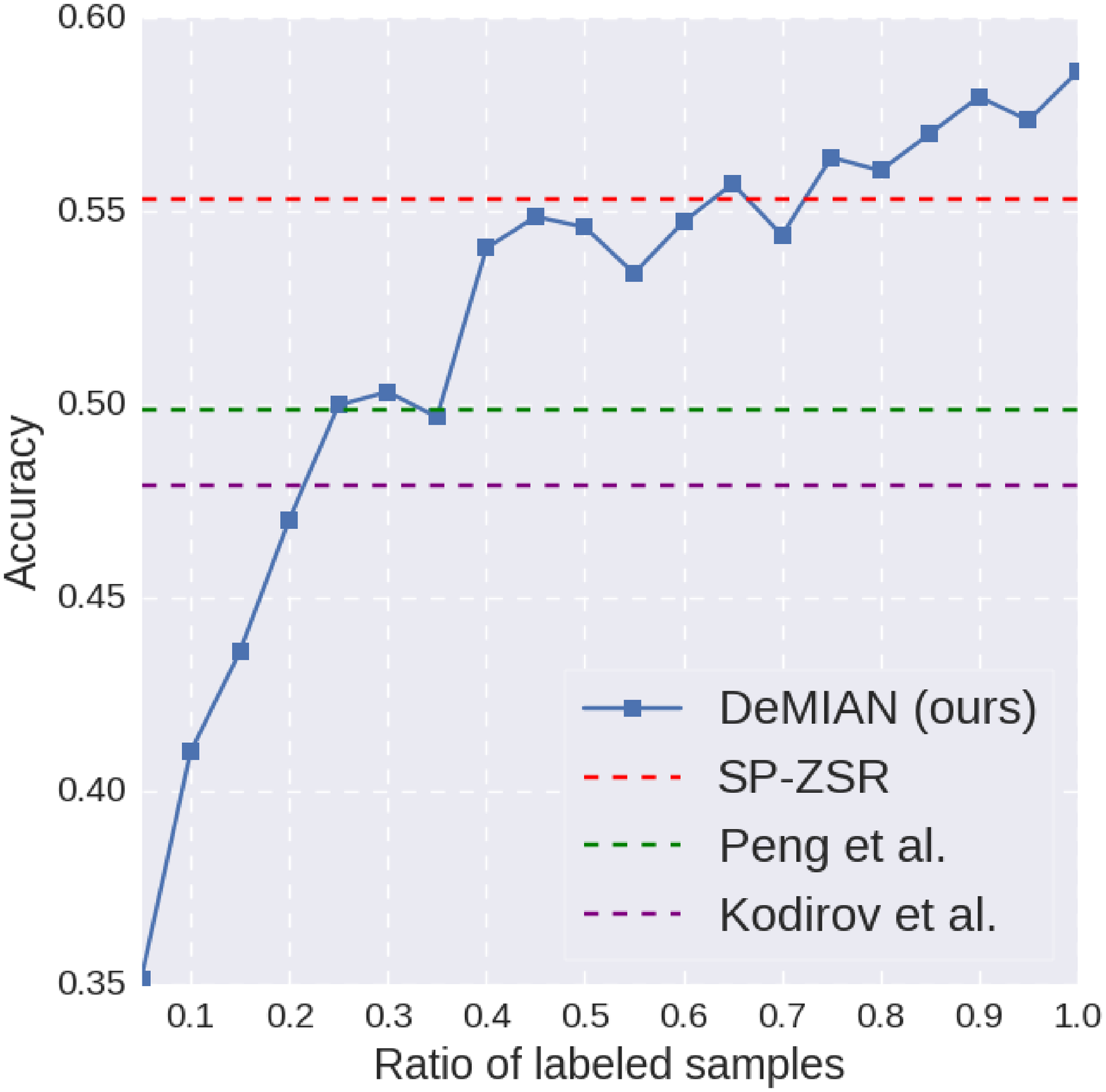}
    \label{fig:mian_increase_cub}}
  \caption{\subref{fig:vgg_feature} and \subref{fig:mian_cub} are the comparison of feature embedding between VGG and our learned representation.  \subref{fig:mian_increase_cub} shows how the zero-shot learning accuracy changed when the number of labeled attribute samples changes.}
 \label{fig:embed_cub}
\end{figure*}
\begin{table*}[t]
\centering
  \begin{tabular}{|c|ccccc|c|} \hline
      Method &cat-dog&plane-auto&auto-deer&deer-ship&cat-truck&Average\\\hline
      Socher \etal\cite{socher2013zero} & 50&65&76&83&{$\bf 90$}&72.8\\\hline
       DeMIAN &{$\bf 60.7$}&{$\bf 74.7$}&{$\bf 86.4$}&{$\bf 88.6$}& 86.9&{\bf 79.5}\\\hline
  \end{tabular}
   \vspace{1mm}
    \caption{Result of zero-shot learning accuracy on CIFAR-10}
  \label{tb:cifar}
\end{table*}
\vspace{-1mm}
\subsection{SUN Attribute}
SUN Attribute consists of images of 717 classes and annotations. For SRL experiment, we used precomputed image features including four kinds of features \cite{xiao2010sun,patterson2012sun}  with a dimension of 19,080. We randomly selected 1,000 dimensions to reduce the dimension. We selected 10 samples for both training and testing per one class. We used the samples of the unseen modality in the training split as the validation data.

The number of units in MIAN (linear) for SRL was  [1000, 300] for the image and [102, 300] for the attribute. The number of units in DeMIAN for SRL was  [1000, 1000, 1000] for the image and [102, 1000, 1000] for the attribute. We used ELU for the activation function of the generator. As a function of $d(G_x{({x_{i}})},G_y{({y_{i}})})$, we used cosine distance and set $\lambda= 10$ for both models. For the discriminator, we used one hidden ReLU layer with the same number of shared hidden units followed by one linear layer. We set learning rate of optimizer to $2.0\times10^{-3}$. A weight decay parameter of $1.0\times10^{-4}$ was used for all layers. We set the mini-batch size to 1,000. 

For zero-shot learning, unlike in the experiment of SRL, we did not use unseen image features for the training. We completely omitted the unseen image features on both the unsupervised training phase and the supervised training phase. We followed the protocol of \cite{zhang2016zero} for image features and splitting datasets. We used the VGGNet \cite{simonyan2014very} features and selected 10 classes for the unseen classes following their settings. We tuned the parameters of our model by using 10 classes of the seen classes. 

The structure used for SUN zero-shot learning was  [4096, 2000, 1000] for the image and [102, 2000, 1000] for the attribute. We used ReLU for the activation function of the generator. As a function of $d(G_x{({x_{i}})},G_y{({y_{i}})})$, cosine distance was used, and we set $\lambda= 10$. For the discriminator, we used three hidden ReLU layers with the same number of shared hidden units followed by one linear layer.  We set learning rate of optimizer to $2.0\times10^{-4}$. A weight decay parameter of $1.0\times10^{-4}$ is used for all layers. We set the mini-batch size to 1,000. 
Then, we reported the 10-fold average of the best score during the supervised training for zero-shot learning.

In the experiment of SRL on SUN, our proposed model showed better accuracy in Table \ref{tb:reg_sun}. The dataset contains only 10 samples for each class; therefore, the random accuracy was less than 1\%, whereas our model showed higher than 4\% for Attribute $\rightarrow$ Image.

We show the result of the zero-shot recognition experiment on SUN in Table \ref{tb:sun_cub}. Our model updated state-of-the-art accuracy about 2\%; notably, our model achieved a state-of-the-art accuracy of using a single method, whereas  the state-of-the-art accuracy was achieved by an ensemble of method \cite{romera2015embarrassingly} + SP-ZSR \cite{zhang2016zeroeccv}. 
\vspace{-1mm}
\subsection{CUB-200-2011}
\vspace{-1mm}
We used the VGGNet \cite{simonyan2014very} features and attributes features following \cite{zhang2016zero}. We used 150 bird species as the seen classes for the training and the remaining left 50 species as the unseen classes following \cite{zhang2016zero} for the testing. We selected 50 seen classes as the validation as in SUN. The number of units of generator was [4096, 1000, 1000] for the image and [312, 1000, 1000] for the attribute. We used ELU \cite{clevert2015fast} for activation of generator. As a function of $d(G_x{({x_{i}})},G_y{({y_{i}})})$, cosine distance was used, and we set $\lambda= 10$. For the discriminator, we used one hidden ReLU layer with the same number of shared hidden units followed by one linear layer. We set learning rate of optimizer to $2.0\times10^{-4}$. A weight decay parameter of $1.0\times10^{-4}$ is used for all layers.  We set the mini-batch size to 1,000. Then, we reported the 10-fold average of the best score during the supervised training for zero-shot learning.

We show the result in Table \ref{tb:sun_cub}. Our model updated state-of-the-art accuracy about 3\%. We compared the representations learned by our network with the VGG features using t-SNE \cite{maaten2008visualizing} in Fig. \ref{fig:vgg_feature} and \ref{fig:mian_cub}. Compared with the VGG-features, we can see that the representations were well separated by their classes. Although some samples were placed far away from the cluster in VGG features, our learned features' distribution was clearly split. We also analyzed our method on a different ratio of the labeled samples. As can be observed from Fig.  \ref{fig:mian_increase_cub}, our method achieved approximately equivalent performance to the current state-of-the-art method \cite{zhang2016zeroeccv} when the ratio was only 0.5.
\vspace{-1mm}
\subsection{CIFAR-10}
\vspace{-2mm}
This dataset consists of 60,000 color images with a resolution of 32$\times$32 pixels (50,000 for the training and 10,000 for the testing) from 10 classes. We followed the settings in \cite{socher2013zero}, in which they used 50-dimension semantic word vectors from Huang's dataset \cite{huang2012improving} that corresponded to each CIFAR category and used an unsupervised feature extraction method \cite{coates2011importance} to obtain 12,800-dimension feature vectors.  They split eight seen classes for training and two unseen classes for the testing. We omitted the unseen classes of the image features during the training. 

The number of units was  [12800, 20, 20] for the image features and [50, 20, 20] for the word vectors. For the discriminator, we used one hidden ReLU layer with the same number of shared hidden units followed by three linear layers. We used ELU for the activation function of generator. we used cosine distance and set $\lambda= 10$. For Discriminator, we used two hidden ReLU layers with the same number of shared hidden units followed by one linear layer. We set learning rate of optimizer as $2.0\times10^{-4}$. A weight decay parameter of $1.0\times10^{-4}$ is used for all layers. We set the mini-batch size to 1,000. In this experiment, we employed a retrieval-based classification method because we can only use two types of word vectors for the testing class. We measured the cosine similarity between the word vectors and the image features. We used the unseen classes in the training split for validation and reported the best accuracy on the basis of validation.

In the experiment on CIFAR-10, our model performed better than the method of Socher \etal\cite{socher2013zero} except for ''cat-truck" as shown in Table \ref{tb:cifar}. 
\vspace{-3mm}
\section{Conclusion}
\vspace{-2mm}
In this paper, we proposed a novel algorithm, named Deep Modal Invariant Adversarial Network (DeMIAN). Our network incorporates the idea of Domain Adaptation and multimodal learning. We aim to learn modality-invariant representations through adversarial training and we could observe the effect of our network in the embedding of learned representation. Our proposed algorithm showed better performance in the experiments of shared representation learning and zero-shot learning. Especially for the zero-shot learning experiment, we achieved state-of-the-art accuracy for CUB-200-2011 and SUN Attribute, which are the benchmark datasets for zero-shot learning. 
\section{Acknowledgement}
This work was funded by ImPACT Program of Council for Science, Technology and Innovation (Cabinet Office, Government of Japan) and  supported by CREST, JST.

{\small
\bibliographystyle{ieee}
\bibliography{egbib}
}

\end{document}